\documentclass[pdflatex,sn-nature]{sn-jnl}

\usepackage{graphicx}%
\usepackage{multirow}%
\usepackage{amsmath,amssymb,amsfonts}%
\usepackage{amsthm}%
\usepackage{mathrsfs}%
\usepackage[title]{appendix}%
\usepackage{xcolor}%
\usepackage{textcomp}%
\usepackage{manyfoot}%
\usepackage{booktabs}%
\usepackage{algorithm}%
\usepackage{algorithmicx}%
\usepackage{algpseudocode}%
\usepackage{listings}%
\usepackage{siunitx}
\usepackage{adjustbox}
\usepackage{makecell}
\usepackage{tikz}
\usetikzlibrary{arrows.meta,positioning,shapes.geometric,fit,calc}

\colorlet{red}{black}

\theoremstyle{thmstyleone}%
\theoremstyle{thmstyletwo}%

\theoremstyle{thmstylethree}%

\begin{document}

\title[Article Title]{Real Time Motion Detection in Video Inspired by Photonic Time Stretch}



\author*[1]{\fnm{Callen} \sur{MacPhee}}\email{cmacphee@ucla.edu}

\author[1]{\fnm{Yiming} \sur{Zhou}}\email{yimingz0416@ucla.edu}

\author*[1]{\fnm{Bahram} \sur{Jalali}}\email{jalali@ucla.edu}

\affil[1]{\orgdiv{ECE Department}, \orgname{UCLA}, \orgaddress{\street{405 Hilgard Ave.}, \city{Los Angeles}, \postcode{90095}, \state{California}, \country{USA}}}


\abstract{Motion perception is fundamental to computer vision and physical AI, and plays a central role in robotics, autonomous vehicles, human–computer interaction, and video compression. Traditional approaches such as background subtraction, frame-differencing and optical flow algorithms often struggle in complex or dynamic scenes, typically sacrificing spatial resolution for computational tractability. In particular, optical flow calculation faces an intrinsic trade-off between full-resolution motion capture, accuracy, and speed, limiting its use in real-time and mission critical applications. Deep learning models offer accuracy but are computationally expensive, not interpretable, and prone to hallucinations when operating outside their training domain. Here we introduce MoViD (Motion estimation via Virtual dispersion and phase Detection), a physics-inspired algorithm grounded in the principles that govern photonic time stretch—a highly successful ultrafast realtime data acquisition platform. Drawing inspiration from the optical physics therein, MoViD treats video as a collection of optical time series waveforms associated with pixels. Motion of an object through a pixel broadens the time series spectrum and, after group delay dispersion, induces a temporal phase that exists only if there is motion. Spatiotemporal group delay dispersion templates, corresponding to dielectric functions, classify specific motion trajectories. {\color{red} Unlike conventional empirical phase-based motion-analysis pipelines that require narrow-band spatial channelization and pyramid reconstruction, the physics-based MoViD operates directly on the broadband video signal, vastly increasing its computational speed. } The resulting algorithm achieves unprecedented performance in both motion characterization and motion magnification, providing full-resolution motion mapping in real time at nearly 200 frames per second at 4K resolution and 70 frames per second for 8K video—orders of magnitude faster than classical optical flow approaches or learning-based methods. Notably, such extreme performance enables real time motion analysis at ultrahigh resolution for motion detection and compensation in robotics and other autonomous devices. Because its formulation is entirely grounded in physics, the algorithm holds promise for implementation in physical hardware.}
\keywords{Motion-Detection, Physical AI, Computer Vision, Photonic Time Stretch, Optical Physics, Physics-Inspired Algorithms}



\maketitle
\section{Introduction}\label{sec1}
The human eye is irresistibly drawn to motion–a survival instinct refined over millions of years, where noticing the {\color{red}faintest flicker could mean the difference between life and death \cite{franconeri2003moving}}. Neuroscience shows that our sense of time is  malleable: danger appears to stretch each instant, while moments of joy seem to vanish in a rush, reminding us that the brain’s internal clock is not a precise metronome but adapts to the rate of change and the information content of events \cite{eagleman2008human}. This deep evolutionary interplay between motion and temporal perception offers a powerful model for machine perception: by attending selectively to change, the brain reduces overwhelming sensory input to what matters most \cite{bruce2009saliency}. 

As such, change detection and motion analysis are cornerstone tasks in computer vision, serving as the foundation for higher-level applications like object tracking, activity recognition, behavior analysis, and video summarization. In practical settings ranging from manufacturing to autonomous vehicles, Physical AI and robotics, the accurate identification of moving objects and dynamic elements is crucial for understanding events in real time. Over the years, various algorithms and models have been proposed to address these tasks, yet they often face challenges in balancing robustness, interpretability, and computational efficiency.

\begin{figure}[!b]
    \centering
    \includegraphics[width=\linewidth]{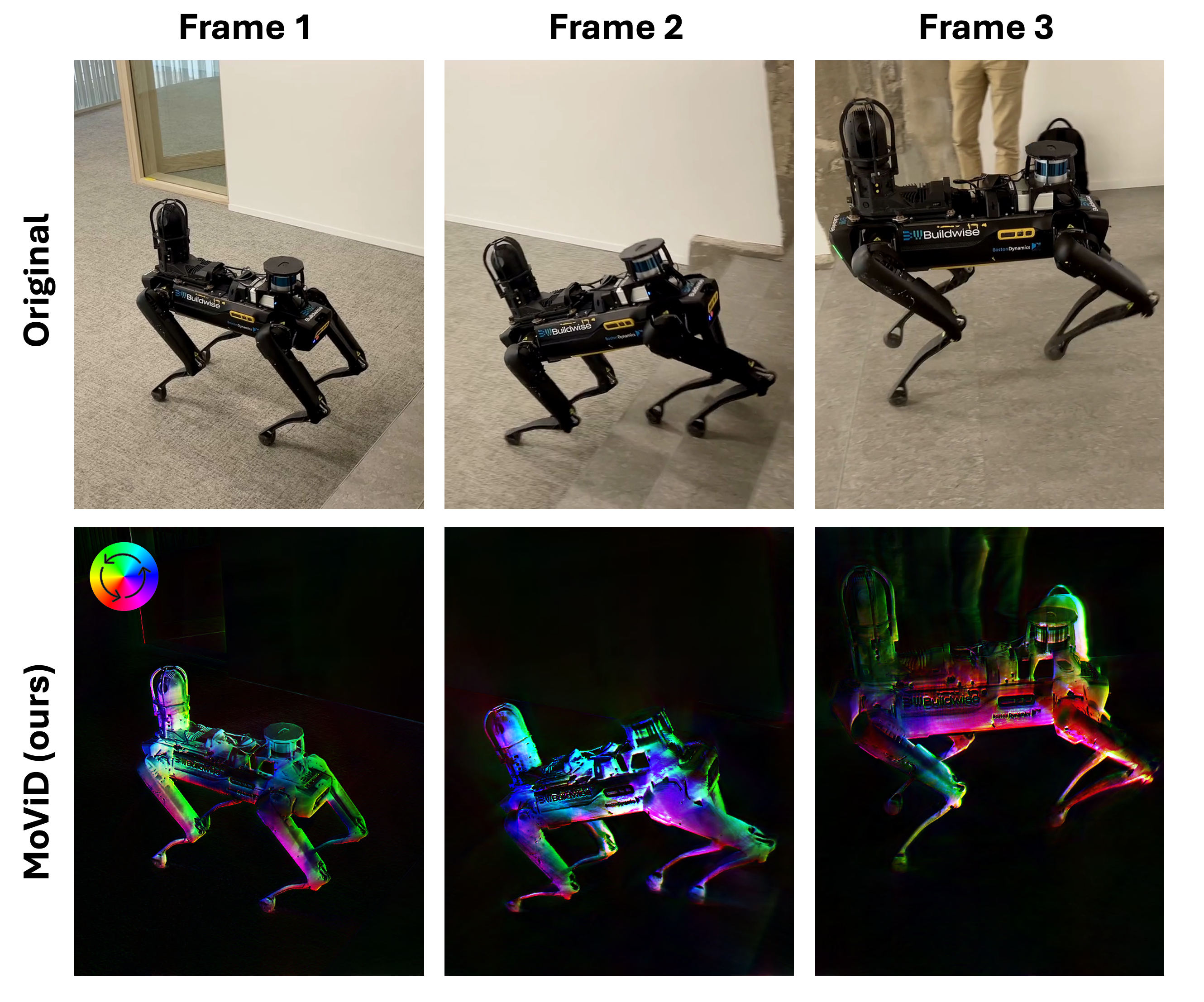}
    \caption{\textbf{Real-Time Full-Resolution Motion Analysis using the MoViD Algorithm.} Three video frames of a quadruped robot (top row) and the corresponding MoViD outputs (bottom row). In the MoViD visualizations, motion directionality is encoded according to the color wheel shown in the bottom-left inset. Notably, visualization of translational and rotational motion is shown as the quadruped rotates and climbs the stairs in the original video. The MoViD algorithm characterizes motion at the pixel-level in full resolution. The resulting motion templates serve as a compact, label-free representation for Physical AI and robotic perception tasks, including anomaly detection, object association, throughput estimation, and long-term system monitoring.
    }
    \label{fig:fig1_teaser}
\end{figure}
\subsection{The Limitations of Traditional and Data-driven Motion Analysis}

A longstanding family of traditional solutions for motion detection relies on rule-based techniques such as background subtraction and temporal differencing ~\cite{horn1981opticalflow}. While effective in controlled environments or with static cameras, these approaches typically require excessive down-sampling and struggle in complex or cluttered scenes where motion is partially occluded or lighting changes rapidly. {\color{red} Such empirical algorithms have also investigated temporal Fourier transforms and spectral representation, often requiring a complex steerable pyramid decomposition to detect and magnify motion in video \cite{wu2012eulerian, wadhwa2013phase, fuchs2010temporal,becker2026fourier}.}

In contrast, the past decade has witnessed significant advances driven by deep learning, where convolutional and recurrent neural networks can directly learn motion features from raw data. Data-driven approaches for optical flow calculation such as FlowNet \cite{dosovitskiy2015flownet} and optical-flow-based CNNs \cite{caldelli2021ofcnn} have achieved state-of-the-art results in action recognition and object detection in videos. RAFT, or Recurrent All-Pairs Field Transforms for Optical Flow \cite{teed2020raft}, represents one widely used approach that has demonstrated exceptional accuracy in optical flow prediction. More recent transformer-based models \cite{huang2022flowformer} have further accelerated vision tasks and contributed to improved performance in video understanding. Additional progress has been made in motion magnification with the adoption of deep learning, with learning-based techniques such as \cite{oh2018deepmag}. However, despite their success, deep learning methods frequently demand huge computational resources and large-scale labeled datasets, limiting their effectiveness in the real world.

\subsection{Physics of Photonic Time Stretch Provides a Blueprint for Motion Analysis}

Emerging in the gap between traditional and data-driven approaches, physics-inspired approaches represent a promising way to reconcile interpretability and performance in algorithm design. Building on the concept of leveraging established physical models as “blueprints” for algorithmic design \cite{phycv2023} \cite{jalali2022symbiosis} \cite{asghari2015pst}, researchers have shown that encoding physical principles into computer vision methods can deliver both computational efficiency and generalizability.

Drawing from fundamental optics and wave propagation principles, we introduce here a novel motion detection, analysis, and magnification technique entitled MoViD (Motion estimation via Virtual dispersion and phase Detection). MoViD represents a completely novel perspective on analyzing and augmenting motion in video. Using simulated spatiotemporal dielectric dispersion to stretch motion in time, MoViD provides an elegant framework to detect, classify, and magnify motion signatures at the pixel-level in real time. As shown in \autoref{fig:fig1_teaser}, MoViD produces dense, directionally encoded motion fields, enabling pixel-level representations suitable for downstream tasks such as anomaly detection, object tracking, and robotic perception.

The remainder of this paper presents the theoretical underpinnings of MoViD and demonstrates how the technique applies optical physics to motion analysis. We first delve into the physical equations that govern chromatic dispersion before outlining how these equations drive robust and interpretable motion detection and classification. We then describe how the approach naturally extends to motion magnification. By bridging established physical theory with modern computer vision needs, MoViD aims to provide a flexible, unified solution for motion analysis across a variety of domains.

\section{Theory}

\subsection{MoViD's Origins in Photonic Time Stretch} 

The physical processes and intuition underlying the MoViD algorithm are rooted in the photonic time stretch technique, a hardware approach for capturing ultrafast, non-repetitive events \cite{jalali1998photonic}. The essence of this method is to modulate time series and other data onto the spectrum of an optical carrier and transform it via spectral phase modulation realized using chromatic (group velocity) dispersion (these are the same steps utilized in the MoViD algorithm, as we will see later). In particular, a dispersion profile with quadratic dependence on frequency produces linear time dilation while higher order dependencies lead to engineered time warping. Over the past two decades, the time stretch framework has become a leading approach to ultrafast single-shot measurements and has spawned numerous variants across spectroscopy, imaging, vibrometry, velocimetry, radar, and lidar \cite{mahjoubfar2017time}. It has enabled the discovery of optical rogue waves; the first real-time observations of soliton-molecule dynamics; insights into the birth of laser mode-locking; direct measurements of relativistic electron beams in accelerators; and the capture of transient chemical dynamics during combustion \cite{mahjoubfar2017time}. Time-stretch imaging extends these capabilities to high-throughput, blur-free quantitative phase imaging and flow cytometry, while time-stretch lidar achieves record-speed four-dimensional (4D) imaging \cite{jiang2020time}. 

MoViD adopts the same principle as photonic time stretch but in the numerical domain. {\color{red}Rather than imposing a spectral phase using chromatic dispersion in the physical domain, MoViD applies a virtual spectral phase in the Fourier domain on the time series representing each pixel’s intensity in the video.} Motion of an object through a pixel broadens the time series spectrum, and group delay dispersion induces a temporal phase that exists only if there is motion. In this way, MoViD mirrors the time-stretch pipeline while remaining lightweight and programmable in the digital domain.

\subsection{Reimagining Video as a 2D Array of Temporally Modulated Waveforms} 	
\begin{figure}[!t]
    \makebox[\textwidth][c]{
        \includegraphics[width=1.1\textwidth]{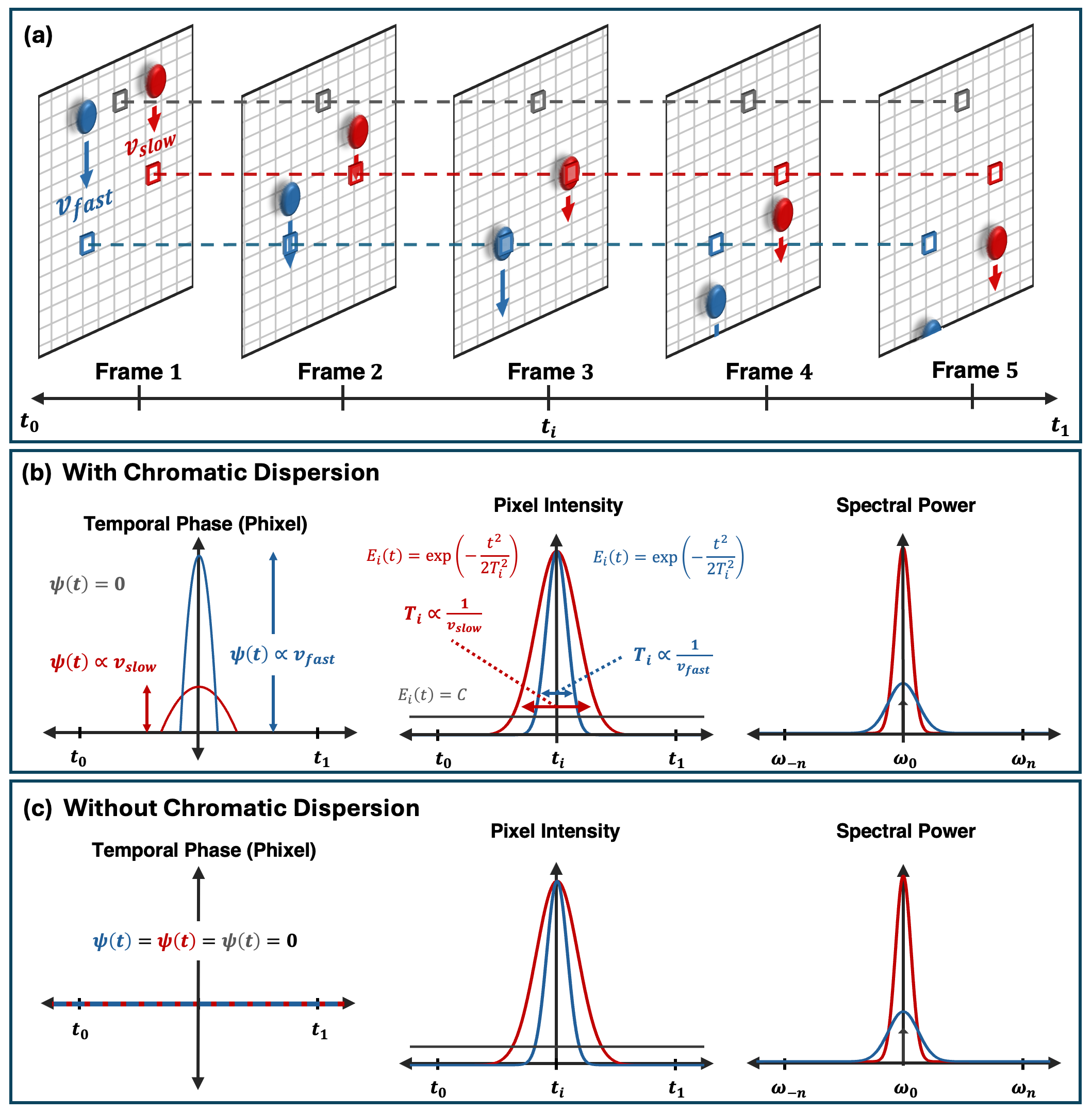}
    }

    \caption{\textbf{Operating and physical principles of MoViD.}
    Together, panels (a--c) illustrate how MoViD converts intensity variations to spectral phase kinematics, enabling pixel-level motion detection.
    \textbf{(a)} Conceptual illustration: two objects that move at different velocities (blue = fast, red = slow) modulate the blue and red pixels time series via their motion. {\color{red}The gray pixel remains static and therefore produces a constant pixel time series.}
    \textbf{(b)} Each time series is modulated via the moving object. Faster motion generates a narrower intensity pulse and broader spectrum. Virtual chromatic dispersion encodes motion in spectral phase. The broader spectrum is stretched more strongly in time, producing a larger temporal-phase modulation $\Delta\psi(t)$; conversely, slower motion yields a narrower bandwidth and a weaker phase response. {\color{red}The constant gray pixel time series has essentially zero temporal bandwidth and therefore produces zero phixel response.} Spectral power remains unchanged.
    \textbf{(c)} Without dispersion, the intensity pulse width, spectral power, and temporal phase are unaltered, so motion cannot be discriminated.}
    \label{fig:fig2_concept}
\end{figure}

Departing from the conventional frame-based perspective for video analysis wherein video is considered a collection of two-dimensional arrays in time, MoViD considers video data as a collection of single-pixel time series arranged across two spatial dimensions. We define a single-pixel time series as the intensity recorded at a specific discrete pixel location $[n,m]$ across time, which we treat as a one-dimensional ``lightfield''. As motion occurs in the broader scene, a single-pixel in space experiences temporal pulses of variation. Variation in a lightfield corresponds to a spectral bandwidth, as the bandwidth is proportional to the magnitude of the change that a time series experiences. This phenomenon is described in \autoref{fig:fig2_concept}. The core conceptual flow is outlined here: 

\begin{enumerate}
\item We re-imagine video as a 2D array of time series signals, each corresponding to a single pixel in the video. 
\item The time series signals are modulated by motion of an object through that pixel, as shown in \autoref{fig:fig2_concept}. The modulation of the time series broadens its spectral bandwidth as dictated by the Fourier transform. 
\item When the time series is subjected to chromatic (group delay) dispersion, it experiences a time-dependent phase, i.e. a chirp. The amount of induced phase is proportional to the spectral bandwidth. 
\item This phase is the signature of motion experienced at that pixel and is at the heart of the proposed motion detection technique. 
\item In the absence of motion, there will be no spectral broadening and hence no temporal phase.
\end{enumerate}

This process transforms the real-valued single-pixel time series into a complex, phase-modulated output. Examining the temporal phase of the dispersed signal provides a measure of the motion occurring at each pixel location over time. We introduce the term “phixel” to describe this temporal phase at each pixel location. This methodology allows for a detailed and nuanced understanding of temporal changes at the \textit{pixel level}. {\color{red}Chromatic dispersion is numerically implemented via a Fourier transform followed by multiplication by a spectral phase kernel.}

Through this process, signals with sharper intensity changes (higher bandwidth) experience more aggressive temporal stretching. Conversely, constant signals undergo no stretching at all. {\color{red}This change-sensitive operation is conceptually analogous to biological visual attention, in which motion and other salient changes receive prioritized processing ~\cite{franconeri2003moving, bruce2009saliency}}. This process reveals a natural mapping between signal change and signal stretch, producing a bandwidth-dependent motion response: the greater the motion in the original signal, the greater the chirp. This fundamental relationship will be mathematically formalized below.

\begin{figure}[!t]
    \makebox[\textwidth][c]{
        \includegraphics[width=0.8\textwidth]{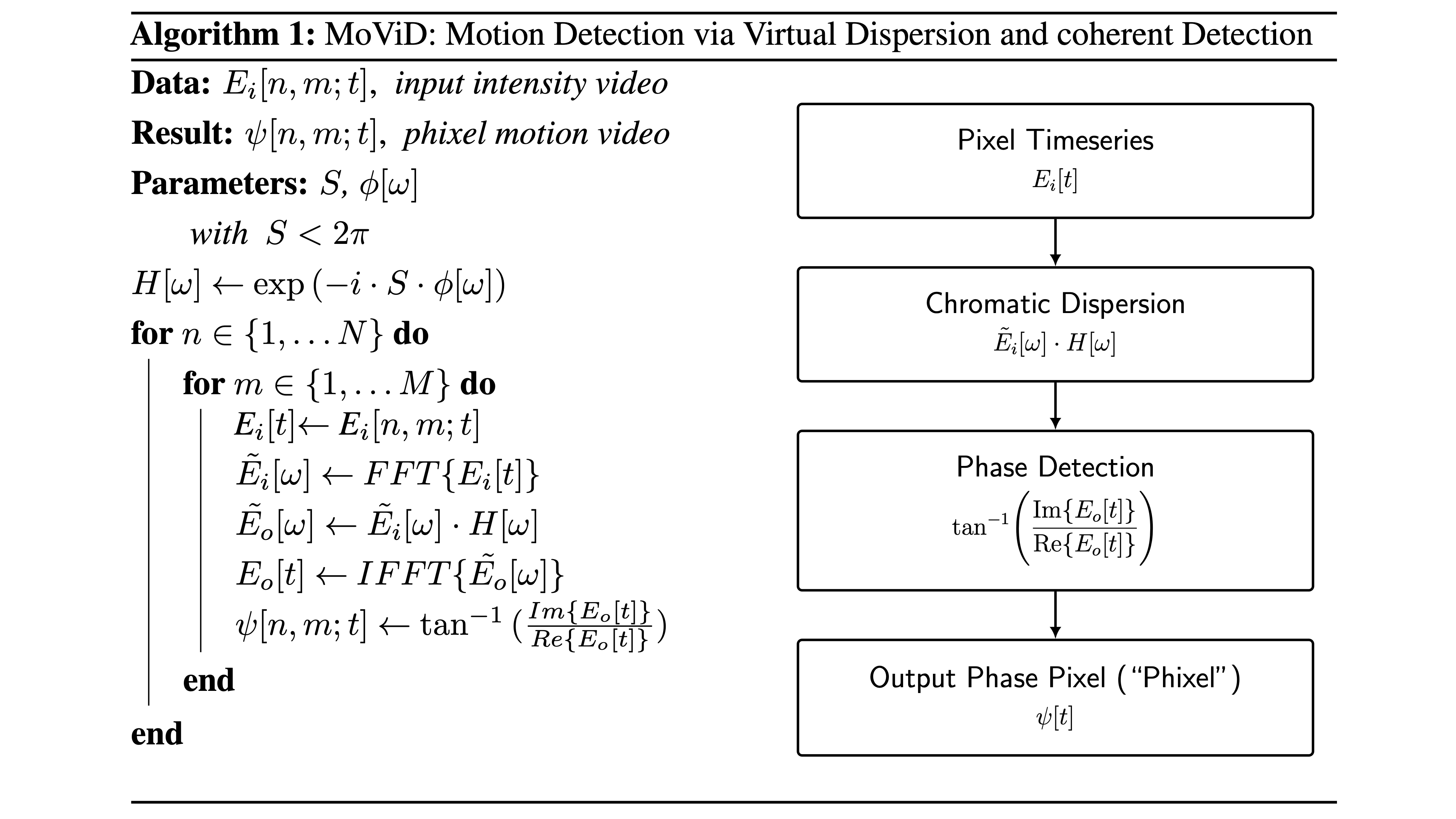}
    }
    \caption{\textbf{Algorithmic Flow of MoViD.} Pseudocode of MoViD and block diagram of the physical pipeline. Each single-pixel time series of the input video is passed through virtual chromatic dispersion, followed by coherent detection to extract the phixel phase map that forms the output motion video. Inputs to the algorithm are a single-channel input video as data and parameters phase strength S and normalized spectral phase kernel $\phi(\omega)$.  Performed independently for every pixel: a 1-D FFT applies a user-defined spectral phase $H[\omega] = e^{-i\phi[\omega]}$; an inverse FFT returns the dispersed signal; coherent phase detection yields the phixel field $\psi[n,m,t]$.}
    \label{fig:fig3}
\end{figure}

\subsection{Group Velocity Dispersion Converts Motion to Chirp} 
 
To formalize the mapping of motion to chirp, we explore the propagation of an initially real-valued Gaussian signal through a linear dispersive medium. In this case, the Gaussian pulse in the pixel time series is caused by motion of an object through that pixel, as described in \autoref{fig:fig2_concept}. For a given 1D Gaussian $E_i (t)$  temporal pulse with standard deviation $T_i$, propagation in the absence of attenuation and nonlinearity is governed by the following partial differential equation: \newline

\begin{equation}
\frac{\partial E}{\partial z}
= i \sum_{m=2}^{\infty} i^m \,\frac{\beta_m}{m!}\,\frac{\partial^m E}{\partial t^m}
\label{eq1}
\end{equation}
\newline
The different order $m$ contributions correspond to dispersive modes \cite{jalali2015tailoring}, which represent the expansion terms in the Taylor series of the mode-propagation constant $\beta$. We can more elegantly describe this transformation as a phase filter in the frequency domain defined by a \textit{spectral phase kernel}

\begin{equation}
\phi(\omega) = \sum_{m=2}^{\infty} \phi_m(\omega) 
= \sum_{m=2}^{\infty} \frac{\beta_m}{m!}\,(\omega - \omega_c)^{m}
\end{equation}
\newline
where $\omega$ represents the frequency with carrier frequency $\omega_c$ and  $H(\omega) = \exp(-i\,\phi(\omega))$ represents the spectral transfer function of the dispersive medium. This is a material property that is defined by the frequency-dependent dielectric function of the medium. Considering only second-order dispersion $\beta_2$, Equation \ref{eq1} is trivially solved in the Fourier domain. After propagation over length $L$, the pulse acquires a temporal phase given by,

\begin{equation}
\psi(t) 
= \tfrac{1}{2} \tan^{-1}\!\Bigl(\frac{L}{L_{D}}\Bigr)
  \;-\;
  \frac{\displaystyle \frac{L}{L_{D}}}{1 + \bigl(\tfrac{L}{L_{D}}\bigr)^{2}}
  \;\frac{t^{2}}{T_{i}^{2}}
\end{equation}
\newline

where $L_D=(T_i^2)/(|\beta_2 |)$ and we assume $\beta_2>0$. Given this temporal phase, the instantaneous frequency, or chirp of the system, is given by

\begin{equation}
\omega(t) 
= \frac{d\psi}{dt}
= -2 
  \,\frac{\displaystyle \frac{L}{L_{D}}}{1 + \bigl(\tfrac{L}{L_{D}}\bigr)^{2}}
  \;\frac{t}{T_{i}^{2}}.
\end{equation}
\newline
The chirp in the pixel time series is highly dependent on $T_i$, which is inversely proportional to the velocity of the object that passed through the pixel. In other words, the chirp is proportional to the object velocity. When $T_i$ approaches infinity, which corresponds to the case of no motion, the chirp vanishes. Conversely, as $T_i$ approaches $0$, indicating ultrafast motion, the chirp grows infinitely large. The output chirp therefore acts as a motion detector, wherein motion is mapped to the instantaneous frequency of the pixel time series. This mapping is controlled by the applied spectral phase profile, whose shape determines how the temporal spectrum is converted into a complex phase response. {\color{red}Concrete kernel examples and their mathematical constraints are provided in the Methods. In particular, the section 6.2.1 of Methods describes important symmetry requirements related to the Hermitian properties.} As discussed later, by designing the dispersive properties we can look for and classify specific spatio-temporal features in the video.  

~\autoref{fig:fig3} presents the algorithmic flow of MoViD for motion detection. Each pixel of the input video is treated independently as a one-dimensional temporal signal and passed through a virtual chromatic-dispersion operator defined by a user-specified spectral phase kernel. Coherent detection of the resulting complex signal yields a pixel-wise phase field that directly encodes motion. This formulation enables dense, full-resolution motion extraction without spatial filtering, training data, or explicit scene modeling, producing a compact motion representation that preserves fine spatial structure.

\subsection{Dielectric Function as a Template for Motion Filtering}

The mapping of motion to temporal phase is governed by the dielectric dispersion function, which serves as a bridge between the temporal phase modulation and the material-like response of the system. Although our prior analysis focuses on the second-order dispersion term \( \beta_2 \) to capture temporal group velocity dispersion (GVD), the general framework naturally admits higher-order dispersive modes that define arbitrary frequency-dependent phase relationships.

{\color{red}In our virtualization scheme, we employ a spectral phase kernel \( \phi(\omega) \) as an intermediary for the equivalent dielectric function of a medium  applied through the phase-only transfer function \(H(\omega)=\exp[-iS\phi(\omega)]\), where \(S\) controls the accumulated phase strength.} For kernels that admit a physical interpretation, the dielectric function can be inverted directly from the chosen spectral phase kernel, providing a direct link between computational filters and their physical analogs. The detailed mathematical derivation of this relationship is provided in the Supplementary Information (SI), where the mapping is shown to follow from equating the spectral phase accumulated through propagation in a dispersive medium with the spectral phase kernel. The resulting expression for the relative permittivity \( \varepsilon_r(\omega) \) is given by

\begin{equation}
    \varepsilon_r(\omega) = 
    \left(
        \frac{S\,c}{L\,\omega}\,\phi(\omega)
    \right)^{2}
    \label{eq:eps_r}
\end{equation}
\newline
where $S$ is the spectral phase strength, a scaling term that determines the strength of the dispersion, $c$ is the speed of light in vacuum, $L$ is the propagation length over which the phase accumulates, $\omega$  is the angular frequency of the temporal signal, and $\phi(\omega)$ is the user-defined spectral phase kernel applied in the frequency domain to achieve motion filtering.

{\color{red}Consequently, the spectral phase kernel serves as the motion-filtering
template in MoViD, while the equivalent dielectric response provides a
physical analogy for kernels that satisfy the relevant material
constraints. Concrete kernel implementations and physical constraints
are described below and in the Methods.}

{\color{red}
\subsection{Spectral Phase Considerations in the Near- and Far-Field}
\label{sec:main_kernel_considerations}

A key point of the MoViD framework is that motion selectivity is not imposed primarily by one particular kernel shape or hand-tuned thresholds. Instead, it follows from the underlying physics of the transformation. The framework therefore works with a broad family of dispersive phase profiles, although the kernel shape influences which temporal or spatiotemporal motion characteristics receive higher attention.

The spectral phase is written as $H[\omega]=\exp(-i\cdot S \cdot \phi[\omega])$, where $S$ is the spectral phase strength. In this work, $S$ distinguishes two nominal operating regimes. For motion detection and classification, we use $S<2\pi$, corresponding to a ``nominal near-field'' regime. In this regime, the virtual dispersion modifies the complex phase of the signal, but the temporal intensity envelope is not meaningfully reshaped. The motion-dependent phase is read out directly as the phixel field with no required unwrapping. Further discussion is provided in ~\cite{jalali2022vevid}.

This differs from conventional (incoherent) photonic time stretch, which typically operates in the temporal far field where stationary phase approximation is satisfied. Here, a stronger dispersion produces a substantial change in the temporal intensity envelope. In this regime, greater accumulated phase allows small motion-induced phase variations to be amplified and visualized. In this far-field operating regime where $S\gg2\pi$, MoViD performs motion magnification, amplifying imperceptible changes in video. Thus, the same virtual-dispersion operator supports phase-domain motion detection at lower $S$ and intensity-domain motion magnification at higher $S$.
}

\subsection{Spatiotemporal Phase Kinematics for Motion Classification}

While the single-pixel perspective yields a mapping of intensity change to phase, incorporation of spatial dependence into our spectral phase kernels generalizes the framework beyond the purely temporal domain. An incoming video signal as a two-dimensional $(x,y)$ lightfield propagating in time $(t)$ has corresponding spatial and temporal frequency values defined by $(k_x  ,k_y, \omega)$. The incoming lightfield is a real-valued sequence of two-dimensional
intensity signals $E_i(x,y,t)$.

When motion occurs within this signal, it manifests as propagation across the transverse spatial axes, $x$ and $y$, of the lightfield. This propagation is a superposition of many spatiotemporal traveling waves, each representing a specific position in Fourier $(k_x, k_y, \omega)$ space. The unique combination of these spectral components results in a distinct motion signature. Much of this motion signature is contained within the spectral phase of the signal, and by modifying and filtering by spectral phase profile we can adjust the \textit{spectral phase kinematics} of the data, thereby detecting or even magnifying motion on the pixel-level. In other words, the dispersion-induced spatiotemporal phase specified by
the applied spectral phase template probes specific patterns of motion
that produce a MoViD output. By defining spectral phase kernels as functions in this space, we can stretch specific portions of the incident signal associated with a given motion class. Our spatiotemporal spectral phase kernels are therefore crafted as \textit{motion templates}, motion feature extractors that are designed to activate, i.e. impart spatiotemporal chirp, for specific classes of motion. 

The chirped output, the complex-valued dispersed signal $E_o (x,y,t)$, contains a spatiotemporal phase produced by the applied motion template. A spatiotemporal phase is produced when the spatiotemporal motion in the video matches that of the applied template. The process decomposes the motion within the lightfield into its primitive spatiotemporal components, enabling efficient filtering and classification for motion characteristics such as direction, magnitude, and periodicity. The explicit angular and speed spectral phase kernels are provided in the Methods.

\subsection{Broadband Nature of Time Stretch Eliminates Channelization and Reconstruction}

\begin{figure}[!t]
    \makebox[\textwidth][c]{
        \includegraphics[width=0.6\textwidth]{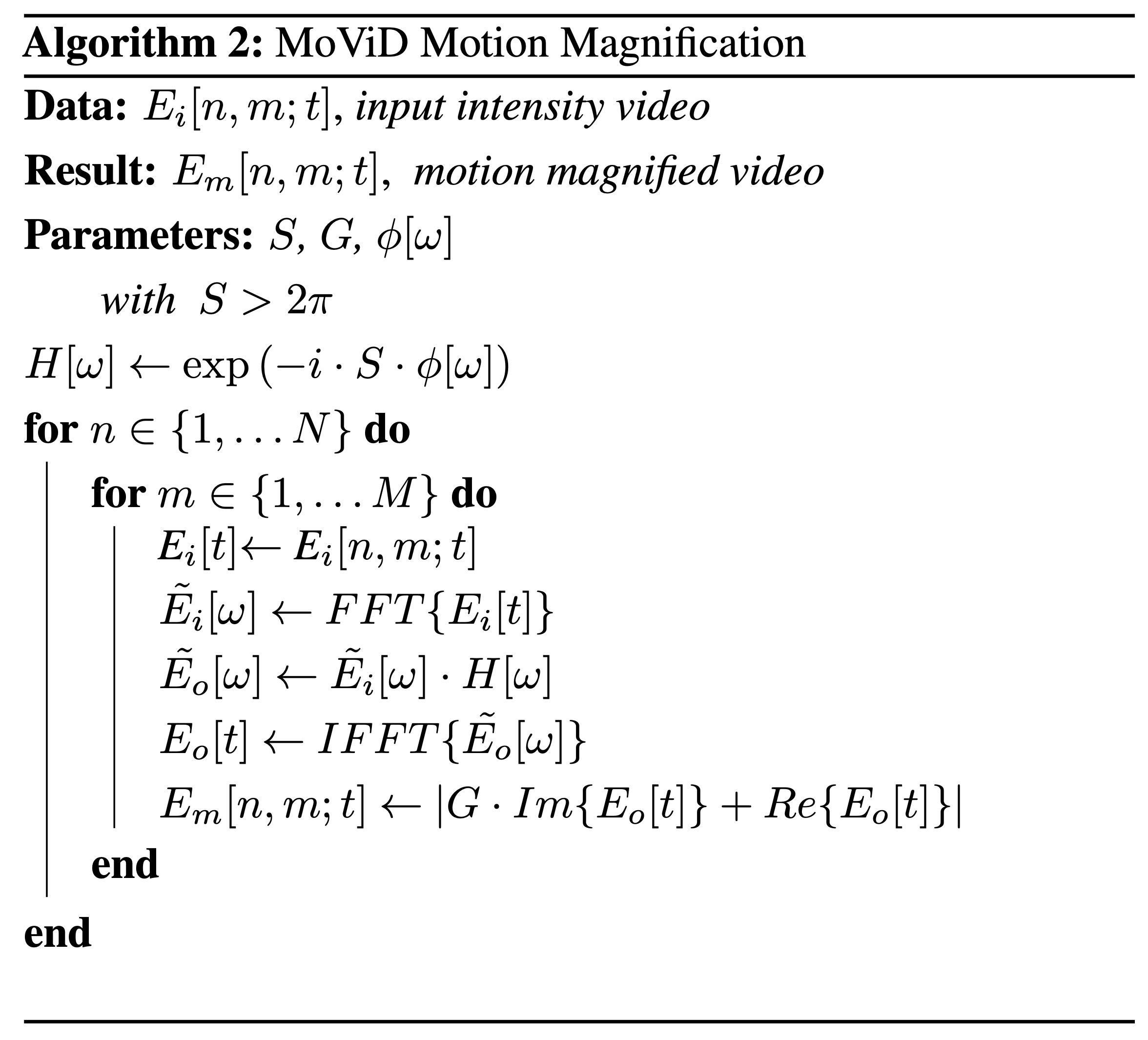}
    }
    \caption{\textbf{Algorithmic Flow of MoViD Motion Magnification.} Pseudocode of MoViD motion magnification. Each single-pixel time series of the input video is passed through virtual chromatic dispersion, followed by magnification of quadrature information. After summation with the in-phase component, the magnitude of the signal serves as the output. Inputs to the algorithm are an input video as data and parameters phase strength S which should be greater than $2\pi$, gain term $G$, and normalized spectral phase kernel $\phi(\omega)$.}
    \label{fig:mag_algo}
\end{figure}

\begin{figure}[!t]
    \centering
    \includegraphics[width=\linewidth]{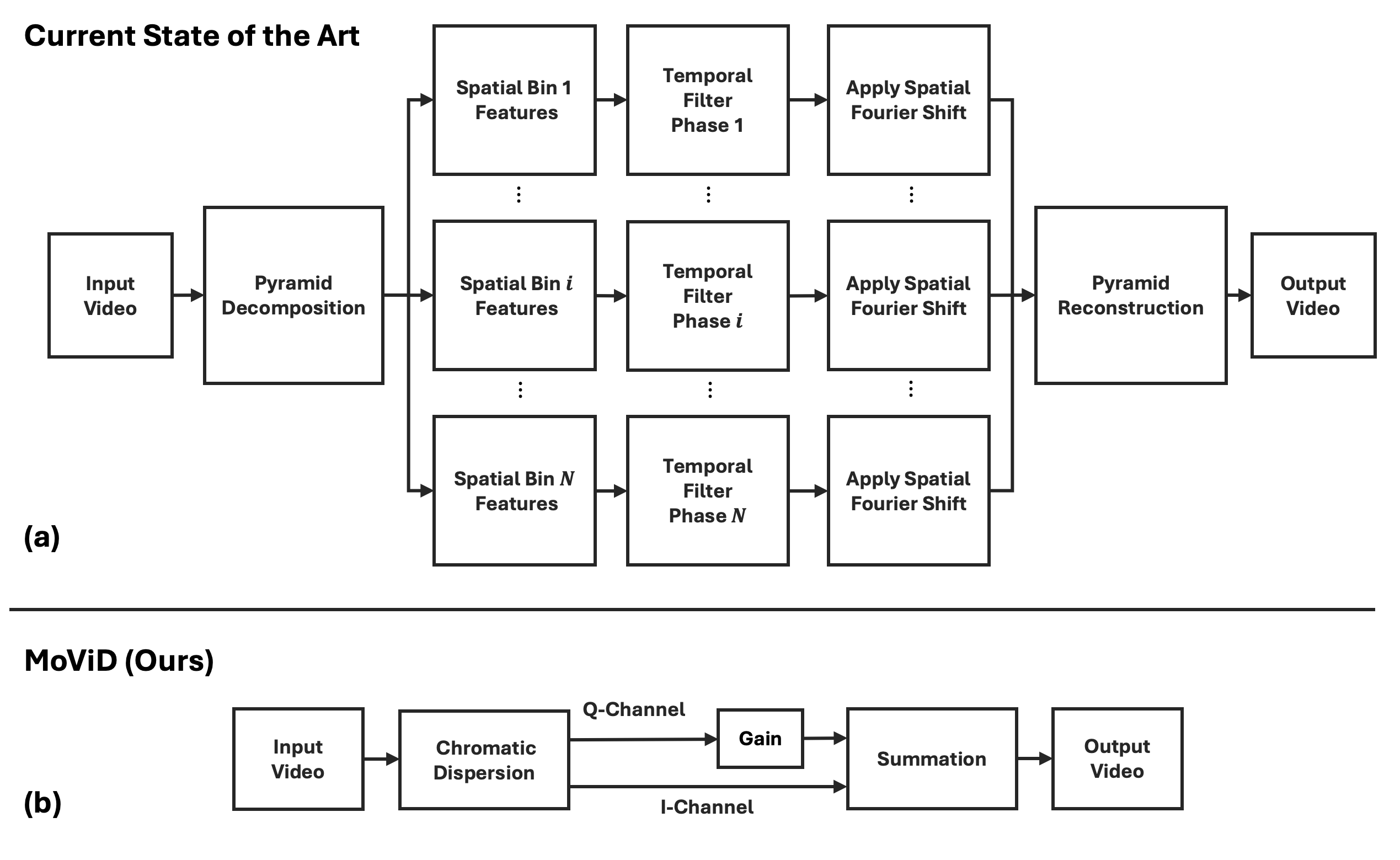}
    \caption{
    \textbf{Algorithmic Block Diagram of MoViD vs. Existing Motion‑Magnification Pipelines.}
    \textbf{(a)} Current state-of-the-art (steerable-pyramid) workflow. Video frames are first decomposed into a multi-scale spatial pyramid; every spatial bin then undergoes phase-specific temporal filtering, spatial Fourier shifting, and a final pyramid reconstruction, resulting in multiple serial passes over the data cube and high memory traffic.
    \textbf{(b)} MoViD processing chain. A single broadband chromatic-dispersion operator generates in-phase (I) and quadrature (Q) channels; coherent gain is applied to the Q-channel and the two streams are summed to produce the output video. Eliminating pyramid decomposition/reconstruction collapses the pipeline to two lightweight stages, greatly reducing computational complexity.
    }
    \label{fig:mag_block}
\end{figure}

Given its basis in spectral phase modulation and measurement, the MoViD algorithm also offers a direct approach for motion magnification – the amplification of minor shifts in video that are imperceptible to the human eye. Traditional phase-based motion magnification methods \cite{wadhwa2013phase} rely on the Fourier shift theorem, which states that a spatial shift in the time domain corresponds to a phase shift in the frequency domain. Such approaches first decompose the input signal using complex steerable pyramids, which break down the image into multiple frequency bands and orientations. This decomposition is necessary because the Fourier shift theorem is fundamentally constrained to single-tone signals, which require the video to be processed in narrow frequency bands. This constraint leads to significant computational overhead, band-limiting artifacts, and reduced spatial resolution.

MoViD introduces a fundamentally different paradigm through its broadband temporal dispersion approach. Rather than requiring complex decomposition into narrow frequency bands, MoViD operates directly on the broadband signal. This approach enables simultaneous processing across all frequencies, preserving the full spectral content of the motion. ~\autoref{fig:mag_algo} illustrates the corresponding MoViD formulation for motion magnification. As in the detection case, each pixel time series is independently dispersed using the same spectral phase framework; however, the quadrature component is selectively amplified before recombination with the in-phase signal. The resulting magnitude encodes motion-magnified intensity while maintaining spatial alignment with the original video. This per-pixel formulation enables controlled motion amplification at full resolution, without spatial pyramids or scale-specific tuning, allowing subtle motions to be enhanced while preserving fine image detail.

~\autoref{fig:mag_block} contrasts the MoViD processing pipeline with conventional motion-magnification approaches based on steerable spatial pyramids. Existing methods require multi-scale decomposition, phase-specific temporal filtering, spatial Fourier shifting, and reconstruction, resulting in multiple computationally expensive serial passes over the video. In contrast, MoViD replaces this workflow with a single virtual chromatic-dispersion operation followed by coherent detection, producing in-phase and quadrature channels that are combined to form the output. By eliminating pyramid decomposition and reconstruction, MoViD collapses the pipeline to two lightweight stages, significantly reducing computational complexity while preserving full spatial resolution. 

By leveraging spectral phase modulation in the virtual dispersion domain, MoViD achieves motion magnification that is extremely computationally efficient while maintaining high spatial and temporal resolution. The elimination of the complex steerable pyramid stage significantly reduces processing requirements, making real-time applications possible even at high resolutions, while still providing precise control over which motion signatures to amplify through tailored spectral phase kernels.

\section{Results}\label{sec2}

\begin{figure}[!t]
    \makebox[\textwidth][c]{
        \includegraphics[width=1.2\textwidth]{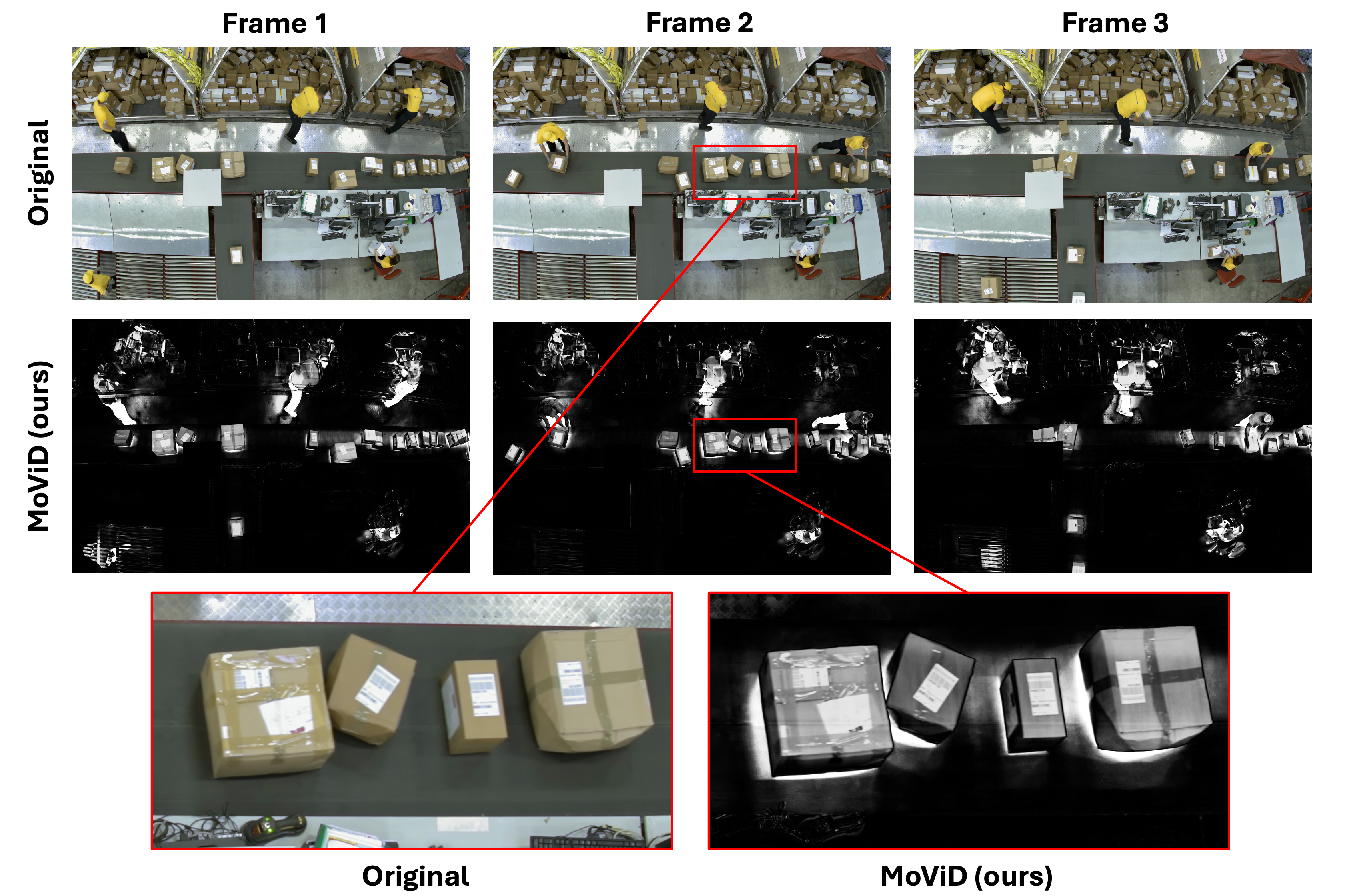}
    }
    \caption{\textbf{Motion Segmentation Capabilities of MoViD in a Real-World Industrial Sorting Scenario.} Three frames from an overhead camera of a manufacturing conveyor sorting system (top row) and the corresponding MoViD outputs (middle row), demonstrating single-pixel-level motion segmentation. MoViD isolates only dynamically changing regions while suppressing any static background, yielding high-contrast representations of moving packages. The highlighted zoom-in (bottom) shows that fine object details are preserved by MoViD, including package edges and printed barcodes, which are clearly visible in the motion domain. This capability enables precise object-level segmentation, sorting, and tracking for industrial inspection, sorting, and automation tasks.}
    \label{fig:metro}
\end{figure}
\subsection{Pixel-Level Motion Segmentation}
A core advantage of the MoViD algorithm is its ability to isolate and represent motion at the resolution of individual pixels while preserving fine-grained structural detail. In \autoref{fig:metro}, we demonstrate MoViD’s performance in a real-world industrial conveyor sorting environment, where only dynamically changing regions are retained and static background structure is suppressed. Despite operating without any training data, MoViD produces high-contrast motion representations of moving packages while preserving subtle object details at the pixel level, including package edges and printed barcodes. This level of detail retention enables precise object-level segmentation and tracking directly in the motion domain, even in visually cluttered scenes. By effectively acting as an intrinsic, physics-driven attention mechanism, MoViD supports downstream tasks such as automated inspection, sorting, throughput estimation, and long-term system monitoring without explicit scene modeling or large labeled datasets.

{\color{red}

To quantify foreground-motion localization, we evaluated MoViD on all four sequences in the baseline category of CDnet2014: PETS2006, Highway, Office, and Pedestrians~\cite{wang2014cdnet}. We compared MoViD with KNN and MOG2 background subtraction as they process standard intensity video without supervised training, semantic object labels, or scene-specific learned models, and are intended for computationally efficient motion localization. As shown in Table~\ref{tab:cdnet_quantitative}, MoViD achieved the highest macro-averaged F1 score and IoU among the methods tested. Details are provided in the Supplemental Information section S1.}

\begin{table*}
\color{red}
\centering
\caption{\textbf{Motion Localization on the CDnet2014 Baseline Category.}
Mean F1 score and IoU are unweighted averages across the four sequence-level
values. Higher values are better. Bold values mark the best result.}
\label{tab:cdnet_quantitative}
\begin{tabular}{lcc}
\toprule
Method & F1 score & IoU \\
\midrule
KNN            & 0.772          & 0.649 \\
MOG2           & 0.688          & 0.557 \\
\textbf{MoViD} & \textbf{0.821} & \textbf{0.707} \\
\bottomrule
\end{tabular}
\end{table*}

\subsection{Classifying Pixel-Level Motion by Direction and Magnitude}

\begin{figure}[!t]
    \makebox[\textwidth][c]{
        \includegraphics[width=1.2\textwidth]{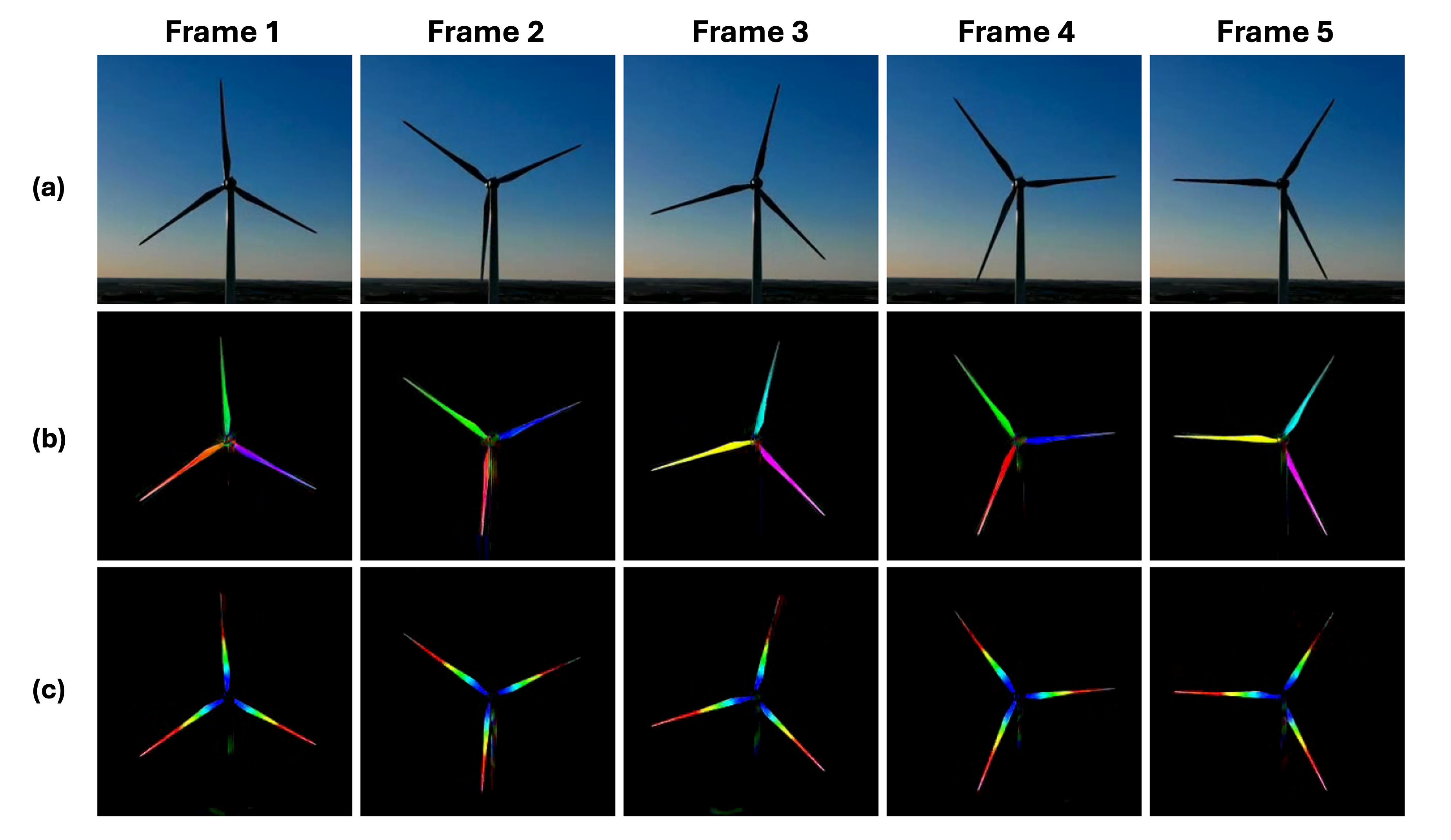}
    }
    \caption{\textbf{Directional and magnitude-based motion classification demonstrated on wind turbine footage}. (a) Original video frames showing turbine rotation. (b) Directional classification of motion components where color represents directionality. (c) Speed-based classification where colors represent different velocity ranges, revealing how blade tips move faster than regions closer to the hub.}
    \label{fig:turbine}
\end{figure}

\autoref{fig:turbine} demonstrates MoViD's ability to segment and classify motion based on user-defined templates encoding specific directions or magnitude ranges. In this example, the middle row contains the color-coded directionality of motion while the bottom row contains the color-coded magnitude of velocity. As the turbine blades rotate, the local motion direction continuously changes, causing the color-coded output to transition in accordance with the prevailing motion orientation. To filter for this directionality, each MoViD motion filter corresponds to a “virtual dielectric” whose dispersive properties target a particular class of spatiotemporal features.

\begin{figure}[!t]
    \makebox[\textwidth][c]{
        \includegraphics[width=\textwidth]{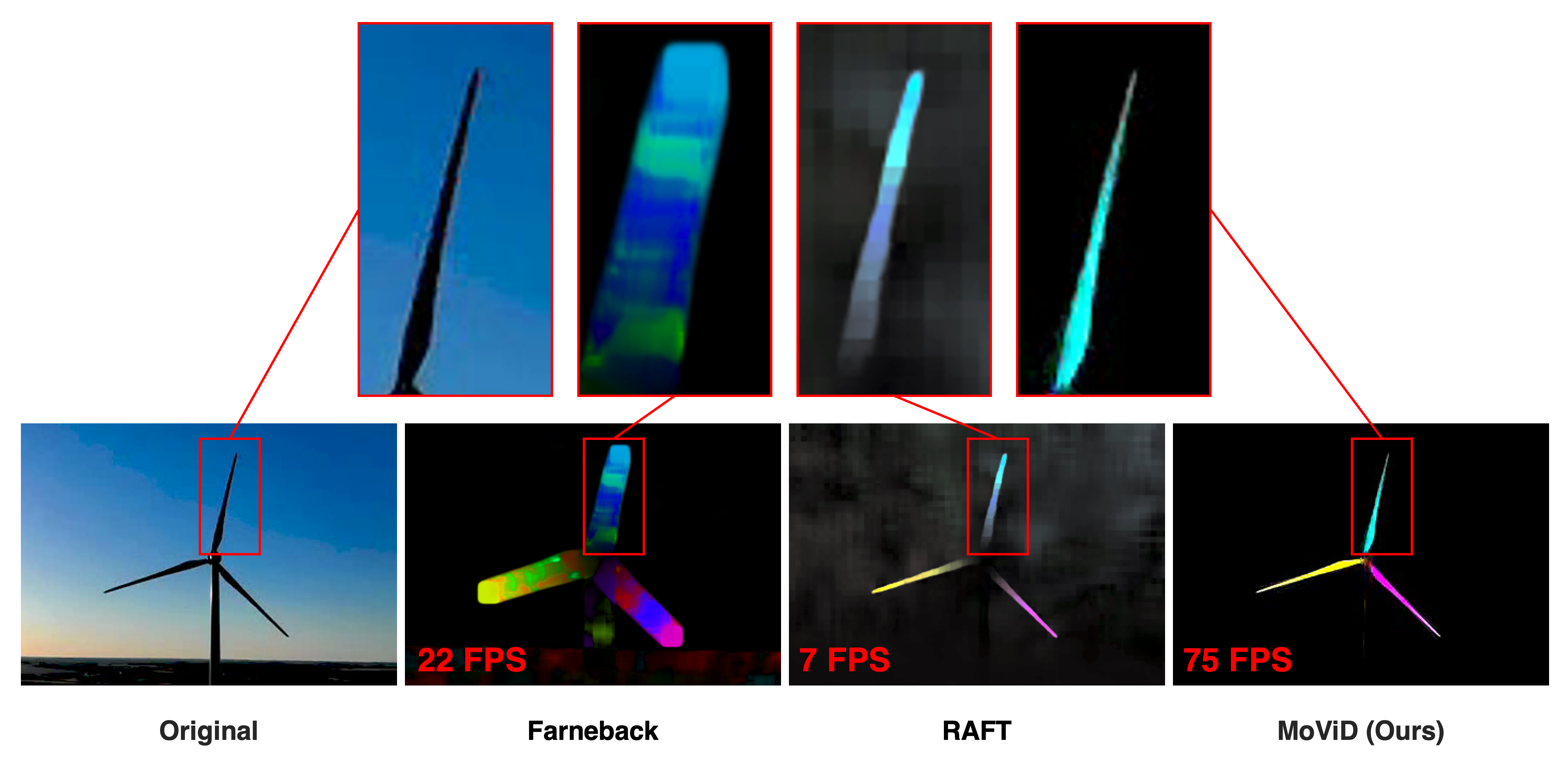}
    }
    \caption{\textbf{Comparison with Traditional and State-of-the-Art Optical Flow}. The left result is generated via the classical Farneb\"ack Method \cite{farneback2003}, the center is generated with the learning-based approach RAFT \cite{teed2020raft}, and the right is MoViD. Whereas the classical approach is too coarse and low-resolution, the learning-based approach introduces artifacts in the background. MoViD shows the highest resolution result. Frame-per-second metrics are shown for each approach, with MoViD running at double the traditional approach and an order of magnitude faster than the deep learning approach.}
    \label{fig:turbine_blowout}
\end{figure}

Beyond direction, MoViD can isolate motion according to magnitude. In the bottom row of  \autoref{fig:turbine}, regions of differing speeds are encoded using distinct color channels. Given the turbine is rotating at a constant angular velocity, the different portions of the turbine blade will travel at different linear velocities given their distances from the axis of rotation. MoViD reveals this difference in linear velocity, encoding the linear speed of the object in color. By designing phase kernels that amplify or attenuate certain ranges of velocity, MoViD becomes a highly selective motion detector capable of parsing out subtle variations within a single scene.

\subsection{Comparative Performance Against Optical Flow}

To highlight MoViD's capability to capture motion details, ~\autoref{fig:turbine_blowout} contrasts two common dense optical flow algorithms with MoViD for the rotating turbine example. The two approaches, the Farneb\"ack algorithm and RAFT, represent two of the most commonly used dense optical flow approaches from the traditional and deep-learning families, respectively. 

The Farneb\"ack approach, while lightweight, relies on extensive downsampling to provide accurate optical flow. In doing so, it loses out on a great deal of detail in high definition scenes and can also exhibit noisy motion outputs, as can be seen in the magnified inset. Alternatively, RAFT as a data-driven approach provides much higher resolution optical flow, but with the added cost of potential hallucination. It can be seen in this example that this approach leads to background artifacts that are not tied to true motion. These hallucinations and artifacts are an added risk in many data-driven approaches. 

In contrast, MoViD yields crisp segmentation and motion boundaries, even for subtle changes, without requiring training or parameter tuning. Regions with minimal motion remain devoid of phase artifacts, whereas regions undergoing large or abrupt motion exhibit correspondingly higher phixel magnitudes. In ~\autoref{fig:turbine_blowout}, MoViD retains the shape of the turbine segment with high fidelity and low noise owing to its pixel-level resolution.

\begin{figure}[!t]
    \centering
    \includegraphics[width=0.8\linewidth]{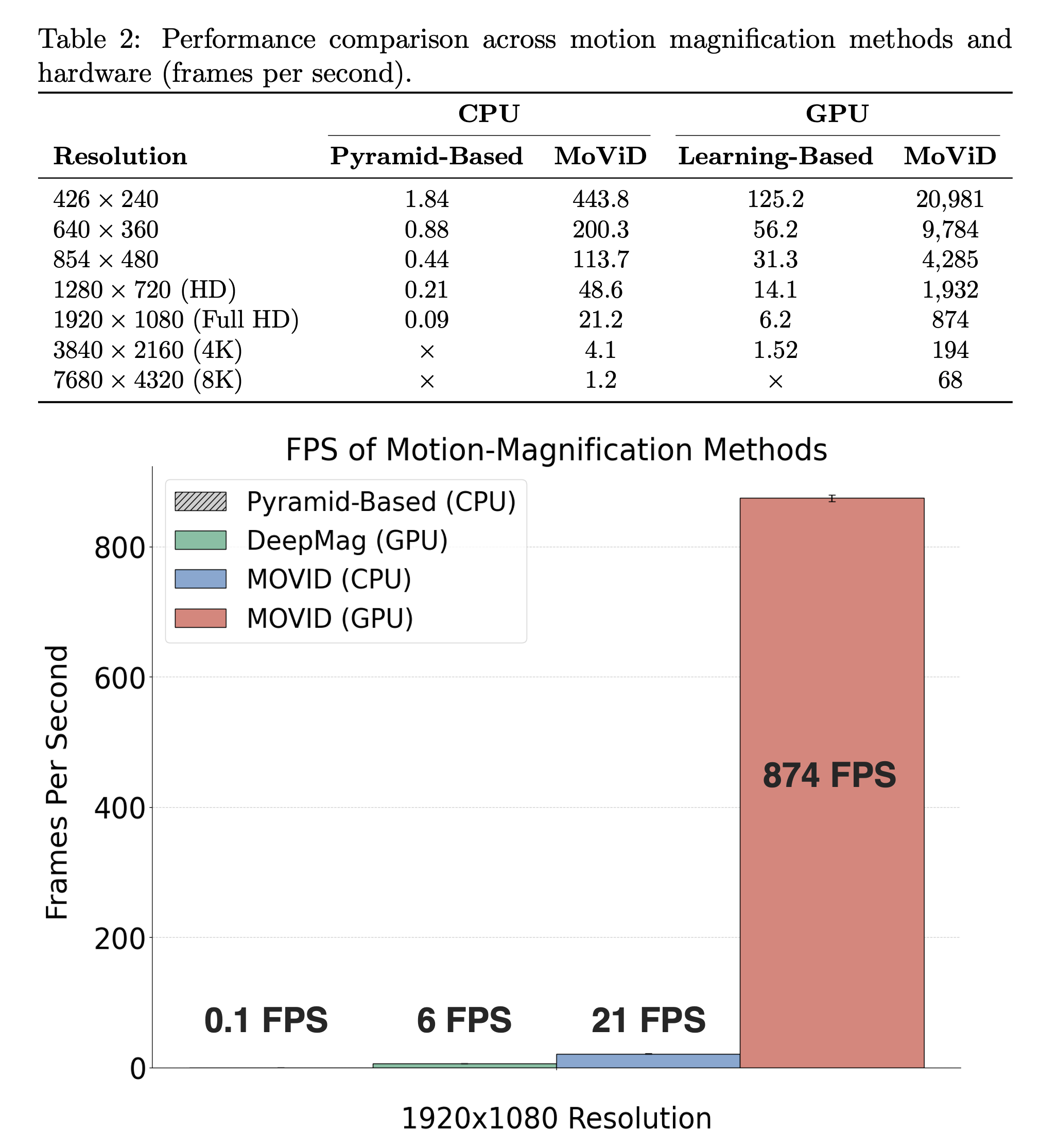}
    \caption{
    \textbf{Runtime Advantage of MoViD over Existing Motion Magnification Pipelines.} (Top) Runtime comparison (FPS) across various resolution scales up to 8K resolution. (Bottom) Visualization of runtime comparison (FPS) at 1080p resolution.
    Throughput benchmarking on an Intel~i9-12900K CPU and NVIDIA TITAN RTX GPU. Across seven resolutions (240p~$\rightarrow$~8K), MoViD vastly outperforms both the steerable-pyramid CPU baseline \cite{wadhwa2013phase} and a learning-based GPU baseline \cite{oh2018deepmag}. At 1080p, MoViD reaches 21~FPS on CPU (vs.~0.09~FPS) and 874~FPS on GPU (vs.~6.2~FPS); at 8K it still sustains 68~FPS on GPU, whereas competing methods fail to achieve real-time performance. These gains exceed two orders of magnitude at every tested resolution and stem directly from the streamlined architecture. Results represented by a $\times$ correspond to the situation in which the algorithm is not able to complete the task on the data given memory or timing constraints of the machine. 
    }
    \label{fig:table}
\end{figure}

\subsection{Real-Time High-Definition Motion Magnification}

A distinctive feature of MoViD is its ability to amplify very small, often imperceptible, motions and vibrations in a scene. By tailoring the dispersion kernel to respond strongly to a specific frequency band corresponding to small displacements or periodic oscillations, MoViD can “stretch” faint signals in time and render them visible. Mathematically, the amplitude of the phase pixel (phixel) is increased. This ``phixel amplification'' is demonstrated in the Supplementary Information videos, where minor vibrations on thin structures (e.g., a swaying tower or a rotating shaft with small-amplitude oscillations) become perceptible.

Notably, owing to its broadband nature, MoViD performs motion magnification in real-time at up to 8K resolution. In comparison, common phase-based \cite{wadhwa2013phase} and deep-learning-based approaches \cite{oh2018deepmag} are orders of magnitude slower. To characterize this speed comparison, we design a dataset composed of six motion amplification videos resized to seven different resolutions. We then compare the runtime of each approach over a collection of frames for each video and average to compute their speed in frames per second. Speed results from a phase-based approach, a deep learning approach, and MoViD are shown in ~\autoref{fig:table}, with comparisons on CPU and GPU hardware. MoViD is over two orders of magnitude faster than the comparative approaches for motion magnification and capable of running at 874 frames per second at 1080p resolution.

{\color{red}
\section{Discussion}

\subsection{Application in Low-latency Motion Analytics in High Resolution Video}

MoViD is well positioned for monitoring and autonomous-perception applications
that require wide-field coverage, fine spatial detail, and high
temporal sampling under latency, bandwidth, or onboard-computation
constraints. This regime arises in civil-infrastructure monitoring,
industrial automation, and robotics, where prior work has demonstrated
video-based vibration measurement, drone-based
displacement sensing, and low-latency visual representations for
resource-constrained navigation
\cite{chen2017video,yang2017blind,ri2024drone,
schoepe2024finding,gehrig2024low}.

MoViD represents temporal intensity modulation as a full-resolution
phixel field. In the conveyor example (\autoref{fig:metro}), this
highlights moving packages and operators while suppressing much of the
static production line. Because the output indicates where motion
occurs rather than assigning semantic identities, it can serve as a
pre-semantic input for container tracking, counting, throughput estimation,
anomaly detection, or updating a digital-twin 
\cite{ward2021real}. With such front-end video processing, subsequent inference can focus on dynamically active regions. In high resolution videos (e.g. 8k) this can unlock significant computation resources for other tasks. 

The motion-magnification mode addresses small structural displacements
that would otherwise go undetected. As shown in the Supplementary
Video, the virtual-dispersion kernel can amplify minute vibrations, motivating use in non-contact screening of bridges, buildings, and industrial machinery. In this setting, MoViD should be
viewed as a motion-localization and visualization framework rather than
a calibrated structural-diagnostic system; deployment will require
validation under camera motion, illumination changes, and environmental loading \cite{kim2024review}.

\subsection{Limitations}

Several limitations follow directly from MoViD's per-pixel formulation. Regions
that are textureless or locally uniform in image coordinates may produce weak phixel responses even when they are semantically part of a moving object. The rotating turbine hub in \autoref{fig:turbine} illustrates this limitation. Near the center, the linear velocity and hence the lateral displacement in the image-plane is small, so the local pixel time series has
little spectral broadening and produces a weak phixel response. Additionally, because MoViD operates in image coordinates, camera motion also contributes to the phixel response. In moving-camera scenes, MoViD measures motion relative to the camera unless global motion is compensated. Established solutions for camera motion compensation include digital image stabilization (DIS), homography- or feature-based registration, camera movement detection using an inertial measurement unit (IMU), or static background anchors \cite{roberto2022survey, guilluy2021video, zitova2003image}.

\subsection{Responsible Use and Dual-Use Considerations}

MoViD is presented here as a motion-amplification and physical motion-analysis method, not as a system for identifying, re-identifying, tracking, or profiling individuals. The present work does not develop, evaluate, or deploy identity recognition, re-identification, biometric classification, person tracking, or behavioral profiling. The reported experiments did not involve any human-subject data collection.

Nonetheless, a fast, full-resolution motion-analysis method could be misused for non-consensual human tracking, surveillance, workplace monitoring, or other privacy-infringing applications. If future applications involve people, they should require appropriate institutional and ethical review and privacy safeguards. When used responsibly, such an ultrafast, high resolution motion analysis technique could unlock a wealth of new applications across the domains of Physical AI, robotics and structural health monitoring.  

}

\subsection{Conclusion}
In this work, we have introduced MoViD, a novel physics-inspired algorithm for motion detection and classification in video sequences. Departing from traditional approaches and the prevailing deep learning paradigm, MoViD takes inspiration from the fundamental physics of temporal group velocity dispersion and coherent detection. By virtualizing these phenomena and mapping them to the digital domain, MoViD offers a unique perspective on motion analysis, interpreting video data as a spatiotemporal lightfield and operating in the spectral phase domain.

The core strength of MoViD lies in its ability to leverage spectral phase kinematics, a qualitatively new representation that captures the intrinsic dynamics of motion at the pixel level. Through the application of tailored spectral phase kernels, or ``motion templates," MoViD can effectively classify motion signatures based on attributes such as direction, magnitude, and higher-order dynamics. This approach not only provides robust and interpretable motion segmentation but also enables hierarchical motion classification, paving the way for a wide range of applications in areas such as object tracking, activity recognition, traffic analysis, and video summarization.

\bibliography{sn-bibliography}

\backmatter

\begin{appendices}




\end{appendices}

\end{document}